%% file: article.tex
\documentclass[runningheads]{llncs}
\usepackage{lineno,hyperref}

\usepackage{multirow}
\usepackage{graphicx}
\usepackage{algorithm}
\usepackage{algorithmicx}
\usepackage{algpseudocode}
\usepackage{float}
\usepackage{lipsum}
\usepackage{algpseudocode}
\usepackage{adjustbox}
\usepackage{txfonts}
\usepackage{tikz}
\usepackage{pgfplots} 
\usepackage{subcaption}
\usepackage{mathptmx}

\usepackage[utf8]{inputenc}
\usepackage[T5]{fontenc}
\usepackage{txfonts}

\begin{document}
\title{Sentence Extraction-Based Machine Reading Comprehension for Vietnamese} 

%
%
\author{Phong Nguyen-Thuan Do\inst{1,2,3} \and Nhat Duy Nguyen\inst{1,2,3} \and Tin Van Huynh\inst{1,2,4} \and Kiet Van Nguyen\inst{1,2,4}\thanks{Corresponding author} \and Anh Gia-Tuan Nguyen\inst{1,2,4} \and Ngan Luu-Thuy Nguyen\inst{1,2,4}}
%
%
\institute{University of Information Technology, Ho Chi Minh, Vietnam \and
Vietnam National University, Ho Chi Minh City, Vietnam \and
\email{\{18520126,18520118\}@gm.uit.edu.vn} \and \email{\{tinhv,kietnv,anhngt,ngannlt\}@uit.edu.vn}}
\authorrunning{Do et al.}
\titlerunning{Sentence Extraction-Based Machine Reading Comprehension for Vietnamese}
%

%
\maketitle              
\begin{abstract}
The development of natural language processing (NLP) in general and machine reading comprehension in particular has attracted the great attention of the research community. In recent years, there are a few datasets for machine reading comprehension tasks in Vietnamese with large sizes, such as UIT-ViQuAD and UIT-ViNewsQA. However, the datasets are not diverse in answers to serve the research. In this paper, we introduce UIT-ViWikiQA, the first dataset for evaluating sentence extraction-based machine reading comprehension in the Vietnamese language. The UIT-ViWikiQA dataset is converted from the UIT-ViQuAD dataset, consisting of comprises 23.074 question-answers based on 5.109 passages of 174 Wikipedia Vietnamese articles. We propose a conversion algorithm to create the dataset for sentence extraction-based machine reading comprehension and three types of approaches for sentence extraction-based machine reading comprehension in Vietnamese. Our experiments show that the best machine model is XLM-R$_{Large}$, which achieves an exact match (EM) of 85.97\% and an F1-score of 88.77\% on our dataset. Besides, we analyze experimental results in terms of the question type in Vietnamese and the effect of context on the performance of the MRC models, thereby showing the challenges from the UIT-ViWikiQA dataset that we propose to the language processing community.

\end{abstract}
\input{section/1-introduction}

\input{section/2-relatedwork}
\input{section/3-ViWikiQADataset}

\input{section/4-Approaches}

\input{section/5-EmpiricalEvaluation}
\input{section/6-Conclusionandfuturework}

%
%
%
\bibliographystyle{splncs04}
\bibliography{references}
%

\end{document}

%% file: section/1-introduction.tex
\section{Introduction}
NLP researchers have studied several works such as question answering, sentiment analysis, and part-of-speech. However, machine reading comprehension (MRC) is one of the critical tasks in NLP in recent years. MRC is an understanding natural language task that requires computers to understand a textual context and then answer questions based on it. MRC is widely used in many applications, such as search engines, intelligent agents, and chatbots.

In the experiments, datasets and models are important in solving the MRC task. For the models, various significant neural network-based approaches have been proposed and made a significant advancement in this research field, such as QANet \cite{DBLP:journals/corr/abs-1804-09541} and BERT \cite{devlin2019bert}. To evaluate these methods, we need different datasets; thus, creating a dataset is vital for computing on MRC models. In recent years, researchers have developed many MRC datasets in other languages such as English \cite{rajpurkar2016squad,yang-etal-2015-wikiqa}, Chinese \cite{cui-etal-2019-span}, French \cite{dhoffschmidt2020fquad}, Russian \cite{Efimov_2020}, Korean \cite{DBLP:journals/corr/abs-1909-07005}, and Italian \cite{SQuADit}.

\begin{table}[H]
\centering
\caption{An example of question-context-answer triples extracted from UIT-ViWikiQA.} 
\label{tab:example}
\resizebox{\columnwidth}{!}{\begin{tabular}{l}
\hline
\begin{tabular}[c]{p{18cm}}\underline{{\bf Context}}:{\bf \textcolor{blue}{Paris nằm ở điểm gặp nhau của các hành trình thương mại đường bộ và đường sông, và là trung tâm của một vùng nông nghiệp giàu có.}} Vào thế kỷ 10, Paris đã là một trong những thành phố chính của Pháp cùng các cung điện hoàng gia, các tu viện và nhà thờ. Từ thế kỷ 12, Paris trở thành một trong những trung tâm của châu Âu về giáo dục và nghệ thuật.Thế kỷ 14, Paris là thành phố quan trọng bậc nhất của Cơ Đốc giáo và trong các thế kỷ 16, 17, đây là nơi diễn ra Cách mạng Pháp cùng nhiều sự kiện lịch sử quan trọng của Pháp và châu Âu. Đến thế kỷ 19 và 20, thành phố trở thành một trong những trung tâm văn hóa của thế giới, thủ đô của nghệ thuật và giải trí. (\it{\textcolor{blue}{Paris is at the meeting point of road and river trade routes and is at the heart of a rich agricultural region.}} In the 10th century, Paris was one of the main cities of France with royal palaces, monasteries, and churches. From the 12th century, Paris became one of Europe's centers for education and the arts. In the 14th century, Paris was the most important city of Christianity. In the 16th and 17th centuries, this was the place where the French Revolution took place and many important historical events of France and Europe. By the 19th and 20th centuries, the city had become one of the world's cultural centers, the capital of art and entertainment.)\end{tabular} \\ \hline
\begin{tabular}[c]{p{18cm}}\underline{{\bf Question 1}}: Vị trí địa lý của Paris có gì đặc biệt? (\textit{
What's so special about Paris's geographical location?})\\ \underline{{\bf Answer}}: {\it{\bf \textcolor{blue}{Paris nằm ở điểm gặp nhau của các hành trình thương mại đường bộ và đường sông, và là trung tâm của một vùng nông nghiệp giàu có.}}} (\textit{\textcolor{blue}{Paris is at the meeting point of road and river trade routes and is at the heart of a rich agricultural region.}})\end{tabular}                                                                                                                                                                                                                                                                                                                                                                                                                                                                                                                                                                                                                                                                                                           \\ \hline
\end{tabular}}
\end{table}

Vietnamese datasets in the MRC task are still really limited over the years. Examples of MRC resources available for Vietnamese language are span-extraction MRC datasets: UIT-ViQuAD \cite{nguyen-etal-2020-vietnamese} and UIT-ViNewsQA \cite{van2020new}, and a multiple-choice dataset: ViMMRC \cite{Nguyen_2020}. Because Vietnamese is a language with insufficient resources for MRC task, we introduce UIT-ViWikiQA, a new dataset for sentence extraction-based MRC for the Vietnamese. Sentence extraction helps readers understand and know more information related to the question than the shorter span. Besides, a new dataset is our vital contribution to assess different MRC models in a low-resource language like Vietnamese. Table \ref{tab:example} shows an example for Vietnamese sentence extraction-based MRC.

In this paper, we have three primary contributions described as follows.
\begin{itemize}
    \item [\textbullet] We build UIT-ViWikiQA, the first dataset for evaluating sentence extraction-based machine reading comprehension for the Vietnamese language, extracted from the UIT-ViQuAD dataset based on our conversion algorithm. Our dataset comprises 23.074 questions on Wikipedia Vietnamese articles. Our dataset is available for research purposes at our website\footnote{\url{https://sites.google.com/uit.edu.vn/kietnv/datasets}}.\\
    
    \item [\textbullet] We analyze the dataset in terms of the question words for each question type (What, When, Who, Why, Where, How, How many, and Others) in Vietnamese, thereby providing insights into the dataset that may facilitate future methods.\\
    
    \item [\textbullet] We propose three types of approaches for the sentence extraction-based MRC for Vietnamese: ranking-based approaches, classification-based approaches, and MRC-based approaches. For each approach, we experiment on different models and compare the experimental results together. In addition, we analyze the dependence of the MRC model on the context.
\end{itemize}


%% file: section/2-relatedwork.tex
\section{Related Work}

The ranking-based approach is inspired by calculating the similarity between two sentences with several previously published algorithms. For example, the word count algorithm is used for the WikiQA dataset \cite{yang-etal-2015-wikiqa}. The BM25 (Best Matching) algorithm is a ranking function used by search engines to rank text according to the relevance of a given query. BM25 is used for text similarity calculation or information retrieval in many previous studies \cite{karpukhin-etal-2020-dense,SARROUTI201796}.

The second approach is the classification-based approach. We use the classification model to find out the sentence containing the answer to the question. The model we use is the maLSTM model \cite{Mueller_Thyagarajan_2016}. maLSTM is a version of the LSTM model whose input is two sentences, and the output is one of two labels (answerable and unanswerable). maLSTM solved the task of the similarity of two sentences \cite{Imtiaz2020DuplicateQP}. In addition, we also use the BiGRU model \cite{cao2019bgru}, which has good performance in classification tasks \cite{9204354,Yin_2021}.

Vietnamese MRC has several datasets to evaluate reading comprehension models, such as the multiple-choice question-answer-passage triple dataset (ViMMRC \cite{Nguyen_2020}) and the span-extraction datasets such as UIT-ViQuAD \cite{nguyen-etal-2020-vietnamese} for Wikipedia-based texts and UIT-ViNewsQA \cite{van2020new} for health-domain news texts. For the MRC models, the first one we use in this paper is the QANet model \cite{DBLP:journals/corr/abs-1804-09541}, which has been applied in previous reading comprehension studies in Vietnamese \cite{nguyen-etal-2020-vietnamese,van2020new}. Next, we use the BERT model and its variants (e.g., XLM-Roberta) proposed by Devlin et al. \cite{devlin2019bert}, and Comeau et al. \cite{con:20}. These models are also used for the previous question answering and MRC tasks in Vietnamese \cite{nguyen-etal-2020-vietnamese,van2020new}, Chinese \cite{cui-etal-2019-span,jing2019bipar}, and English \cite{devlin2019bert,Qu_2019}. We also use a variant of BERT called PhoBERT as a monolingual pre-trained language model \cite{phobert} for Vietnamese.

%% file: section/3-ViWikiQADataset.tex
\section{The ViWikiQA Dataset}
\label{viwikiqa}
We describe the UIT-ViWikiQA dataset and the process of creating it from the UIT-ViQuAD dataset (proposed by Nguyen et al. \cite{nguyen-etal-2020-vietnamese}), as well as several analyses on our dataset.

\subsection{Task Definition}
The main task in this paper is \textbf{sentence extraction-based machine reading comprehension} for Vietnamese. This is a task that requires the computer to find the sentence in a context that contains the answer to a specific question. The input to the task includes question Q and the context S = $(S_1, S_2...S_n)$ containing the answer. The output of the task is a sentence $S_i$ in the context, the condition that this sentence $S_i$ includes the answer to the question Q in the input. The examples we show are in Table \ref{tab:example}, where the input consists of \textbf{Question} and \textbf{Context}, the output is \textbf{Answer}, the sentence is colored in the \textbf{Context}.

\subsection{Dataset Conversion}

We create the UIT-ViWikiQA dataset based on the UIT-ViQuAD dataset. Answers and their answer start are updated by Algorithm \ref{al:conversion}.
\begin{algorithm}
\caption{Converting the UIT-ViQuAD dataset into the UIT-ViWikiQA dataset.}\label{al:conversion}
\textbf{Input:} {The context {\bf C} and the answer start of answer {\bf AS}}\\
\textbf{Output:} {Returning the sentence {\bf ST} that contains the answer and its start {\bf SS} }.
\begin{algorithmic}[1]
\Procedure{Converting the dataset}{}
\State{Sentences $\leftarrow$ Segmenting the context C into the list of sentences};
\State{Start $\leftarrow$ 0};
\State{End $\leftarrow$ -1};
\For{S in Sentences}
    \State{Start $\leftarrow$ End + 1};
    \State{End $\leftarrow$ Start + len(S)};
    \If{$Start \leq {AS}< End$} 
    \State{$break$};
    \EndIf
\EndFor
\State{ST $\leftarrow$ i};
\State{SS $\leftarrow$ C.find(ST); //SS is starting position of sentence ST in context C.}
\State \textbf{return} $ST,SS$
\EndProcedure
\end{algorithmic}
\end{algorithm}

In the UIT-ViQuAD dataset, a sample includes context, question, answer, answer start, and id. Based on {\bf Algorithm 1}, we change the answer and the answer start using context and answer start {\bf (AS)} of UIT-ViQuAD. Next, we use a sentence segmentation tool to segment the context {\bf (C)} into the list of sentences $S=(S_1,S_2,S_3,...S_n)$, then $S_{i}$ $\in$ S is an answer of the question in UIT-ViWikiQA such that: $Start(S_i) \leq{{\bf AS}} < End(S_i)$, where Start($S_i$) and End($S_i$) are the starting and ending positions of the $S_i$ in context.

\subsection{Dataset Analysis}

Similar to UIT-ViQuAD, UIT-ViWikiQA comprises 23,074 questions based on 5,109 passages collected from 174 articles. We split our dataset into three sets: training (Train), development (Dev), and Test. Table \ref{tab:overallstatistics} shows statistics on the number of articles, passages, and questions in each set.

\begin{table}
\centering
\caption{Overview of the UIT-ViWikiQA dataset.}
\label{tab:overallstatistics}
\begin{tabular}{l|r|rrr}
\hline
\multicolumn{1}{c}{\textbf{}} & \multicolumn{1}{|c|}{\textbf{All}}& \multicolumn{1}{c}{\textbf{Training set}} & \multicolumn{1}{c}{\textbf{Development set}} & \multicolumn{1}{c}{\textbf{Test set}} \\ \hline
Articles & 174 & 138 (79.32\%) & 18 (10,34\%) & 18 (10,34\%) \\ 
Passages & 5,109 & 4,101 (80.27\%) & 515 (10.08\%) & 493 (9.65\%) \\ 
Questions & 23,074 & 18,579 (80.52\%) & 2,285 (9.90\%) & 2,210 (9.58\%) \\ \hline
\end{tabular}
\end{table}

Based on UIT-ViQuAD proposed by Nguyen et al. \cite{nguyen-etal-2020-vietnamese}, we divide the questions of the UIT-ViWikiQA dataset into seven categories: What, When, Who, Why, Where, How, How many\footnote{Questions are related to quantities, numbers, and ratios.}, and Others. The Vietnamese question types were done manually and inherited from the UIT-ViQuAD dataset.

\begin{figure}[H]
    \centering
    \includegraphics[width=0.75\linewidth]{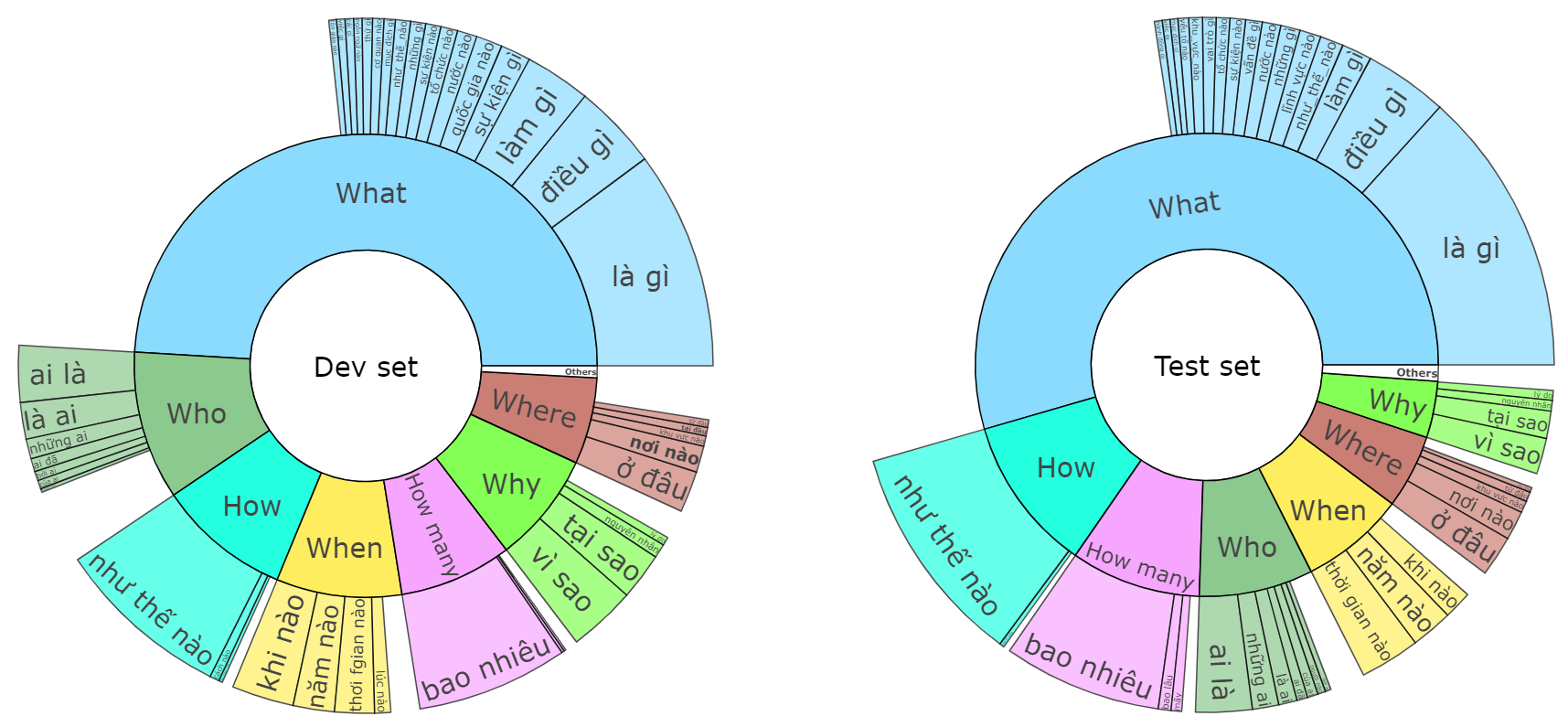}
    \caption{Distribution of the question types on the Dev and Test sets of UIT-ViWikiQA.}
    \label{fig:question_types}
\end{figure}

Figure \ref{fig:question_types} shows the distributions of question types on the Dev and Test sets of our dataset. What questions account for the largest proportion, with 49.06\% on the Dev set and 54.48\% on the Test set of our dataset. The type of questions with the lowest proportion is the Others with 0.83\% on the Dev set and 1.13\% on the Test set. The remaining types of questions account for a fairly similar proportion from 4\% to 10\% in both Dev and Test sets.

In Vietnamese, the question words to ask for each type of question are varied, as shown in Figure \ref{fig:question_types}. The What question has the most range of question words. Figure \ref{fig:question_types} only shows the words to question with a large frequency of occurrence (approximately 0.7\% or more). For example, in What questions, question words such as "là gì", "cái gì", "điều gì", or "làm gì" all mean "what" where "là gì" has the highest rate with 20.69\% in the dev set and 24.42\% in the test set. Similar to What questions, the phenomenon of varied words to question also occurs in the Who, When, Why, and Where question types. Specifically, for the Vietnamese When question type, the question words to pose such as "khi nào", "năm nào" or "thời gian nào" has the same meaning as the question word "When". However, the How question type and the How many question type, the question words are mostly the same. The question word "như thế nào" accounts for 87.79\% of the How questions in the Dev set and 95.02\% in the Test set. In the UIT-ViWikiQA dataset, it is difficult for the models to not only in deductive questions such as Why and How but also in the variety of question words in each question type.

%% file: section/4-Approaches.tex
\section{Approaches}
\label{approachs}
In this section, we propose three different approaches to sentence extraction-based MRC for Vietnamese.
\subsection{Ranking-Based Approaches}

\label{ranking_appraoch}

The ranking approach is inspired by calculating the similarity of two sentences. From the context, we choose a sentence that has the greatest similarity with the question, which means that the selected sentence contains the answer to the question. For example, given a question Q and $S=(S_1, S_2, S_3,...,S_n)$ are sentences in a context containing the answer to question Q. The sentence ranking task is that $S_i$ is assigned a value $rank(S_i)$ for each sentence, which is the similarity between question Q and sentence $S_i$. For each pair of sentences $S_i$ and $S_j$ if $rank(S_i) > rank(S_j)$, the answer to question Q is more likely to be in question $S_i$ than in question $S_j$ and vice versa. 
\begin{figure}[H]
    \centering
    \includegraphics[width=0.75\linewidth]{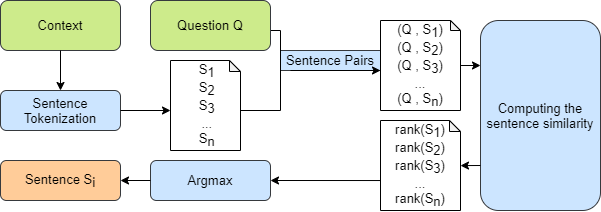}
    \caption{The overview architecture of the sentence ranking-based approach.}
    \label{fig:ranking_appraoch}
\end{figure}

Figure \ref{fig:ranking_appraoch} describes the process of solving the sentence extraction problem following the sentence ranking approach. Input includes context and question Q; after tokenizing sentences for context, we obtain a set of sentences $S= ( S_1, S_2, S_3,..,S_n )$. Then, for each sentence $S_i$ $\in$ S is paired with Q, each $(Q, S_i)$ is calculated a value representing the similarity between $S_i$ and Q. The result is $S_i$, under the condition that $S_i$ is the one with the greatest similarity to the question Q among the sentences in the set S.

\subsection{Classification-Based Approaches}

\begin{figure}
    \centering
    \includegraphics[width=0.75\linewidth]{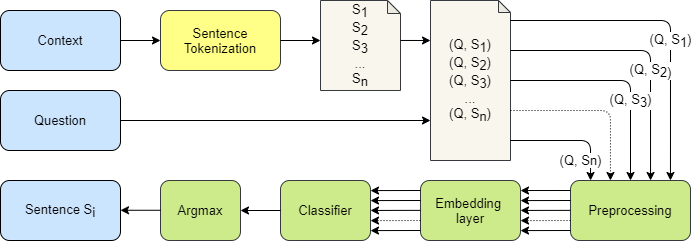}
    \caption{The overview architecture of the text classification-based approach.}
    \label{fig:classification_approach}
\end{figure}

In the second approach, we apply the classification model to solve the task of sentence extraction-based MRC. For example, give a question Q and $S=(S_1, S_2, S_3,.. S_n)$ are sentences in a context that houses the answer for question Q, then we pair each $S_i \in S$ with question Q into n pairs of sentences $(Q, S_i)$. We conduct preprocessing n pairs of sentences to match the model. For each pair $(Q, S_i)$, we use the binary classification model with two labels, 0 and 1. If the label of the sentence pair $(Q, S_i)$ is 1, it means that $S_i$ is the sentence containing the answer to question Q, and if the question pair $(Q, S_i)$ is labeled 0, that means sentence $S_i$ is unanswerable for the question Q.

Figure \ref{fig:classification_approach} depicts the process of solving the sentence extraction problem based on a classification approach. Similar to \ref{ranking_appraoch}, input includes question Q and a context contains the answer to question Q. Next, separate the sentences for the context and obtain a set of sentences $S=(S_1, S_2, S_3,.. S_n)$, each sentence $S_i \in S$ are combined with Q and get n sentence pairs $(Q, S_i)$. N sentence pairs $(Q, S_i)$ are preprocessed by algorithms suitable for the model type. Then we use the trained word embedding models to encode text into vectors of numbers. The binary classification model takes the newly coded vectors as input to define the label for each data sample. The argmax function ensures that the context has only one sentence labeled 1. The output receives a sentence $S_i$ that contains the answer to the question Q.

\subsection{Machine Reading Comprehension-Based Approaches}

\begin{figure}
    \centering
    \includegraphics[width=0.75\linewidth]{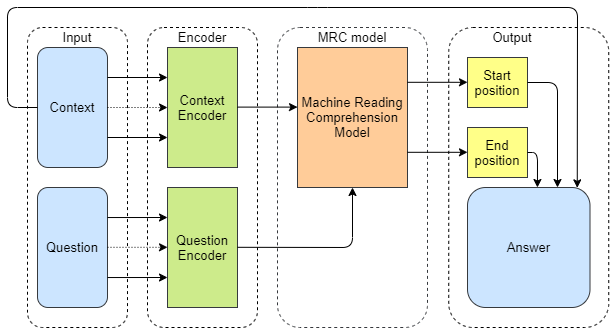}
    \caption{The overview architecture of the MRC-based approach.}
    \label{fig:mrc_approach}
\end{figure}

We define the input and output for the task of reading comprehension. Given a question Q and a context $S=(S_1, S_2, S_3,.. S_n)$ consisting n sentences. The goal of the model is to identify a sentence $S_i \in S$ that answers question Q.

Figure \ref{fig:mrc_approach} depicts the architecture of our approach. The model consists of four components: input, encoder, machine reading comprehension model, and output. Inputs include a question Q, and a context S provides the answer to question Q. Next, the encoder converts the context and question into vectors of numbers with a format that matches the type of MRC model in the next step. The machine reading comprehension model takes the vectors of numbers as input and processes it to give the start and end position of the answer in the context S. With the start and end positions of the answers we just found from the model, we combine them with context to find the sentence $S_i \in S$ for answers question Q in the input.

%% file: section/5-EmpiricalEvaluation.tex
\section{Experiments and Result Analysis}

\subsection{Experimental Settings}

For the sentence ranking approach, we use the underthesea toolkit\footnote{https://pypi.org/project/underthesea/} to separate the context into individual sentences. Based on Vietnamese language characteristics, we utilize Python Vietnamese Toolkit\footnote{https://pypi.org/project/pyvi/} to segment words. We implement the word count algorithm and the BM25 algorithm to evaluate. For BM25, we use Rank-BM25: a two-line search engine\footnote{https://pypi.org/project/rank-bm25/} to aid in computation.

For all models of  sentence  extraction-based  machine  reading  comprehension using the text classification-based approach, we use a single NVIDIA Tesla K80 via Google Colaboratory to train them. Based on the word structure of the Vietnamese language, we use VnCoreNLP\footnote{https://github.com/vncorenlp/VnCoreNLP}, published by Vu et al. \cite{vu-etal-2018-vncorenlp}, to separate words. We use the pre-trained word embedding PhoW2V introduced by Nguyen et al. \cite{phow2v_vitext2sql} for maLSTM and BiGRU. We set batch size = 64 and epochs = 20 for both two models.

In the MRC-based approach, we train models in a single NVIDIA Tesla K80 via Google Colaboratory. We employ PhoW2V \cite{phow2v_vitext2sql} word embedding as the pre-trained Vietnamese word embedding for the QANet \cite{DBLP:journals/corr/abs-1804-09541} model to evaluate our dataset, and we set epochs = 20 and batch-size = 32. We use the baseline configuration provided by HuggingFace\footnote{https://huggingface.co/} to fine-tune a multilingual pre-trained model mBERT \cite{devlin2019bert}, the pre-trained cross-lingual models XLM-R\cite{con:20}, and the pre-trained language models for Vietnamese PhoBERT \cite{phobert}. We set epochs = 2, learning rate = $2e^{-5}$, a maximum string length of 384, and maximum query length of 128 for all three models. For the characteristics of the PhoBERT$_{Base}$ model, we follow its default parameters except for the maximum string length which is set to 256 and use VnCoreNLP to segment the word for data before training with PhoBERT.


\subsection{Evaluation Metrics}

To evaluate performances of MRC models, we use two evaluation metrics following UIT-ViQuAD \cite{nguyen-etal-2020-vietnamese} and SQuAD \cite{rajpurkar2016squad}, including Exact Match and F1-score. Considering a predicted answer sentence and a human-annotated answer sentence, if the two sentences are exactly the same, EM is set to 1, EM is set to 0 otherwise. F1-score measures the overlap tokens between the predicted answer and the human-annotated answer. 

\subsection{Experimental Results}

Table \ref{tab:resultmodel} shows the performance of models on the Dev and Test sets. The models in the classification-based approach achieve the lowest performance, and the highest performance belongs to the models of the MRC-based approach. The best model (XLM-R$_{Large}$) reaches 88.77\% (in F1-score) and 85.87\% (in EM) on the Test set. The performance of the best model in the ranking-based approach (BM25) with 10.54\% in F1-score and 12.70\% in EM. When compared to the best model of classification-based approach (BiGRU), the XLM-R$_{Large}$ model outperforms with 26.21\% (in F1-score) and 30.71\% (in EM).

\begin{table}[H]
\centering
\caption{Machine performances (EM and F1-score) on the Dev and Test sets.}
\label{tab:resultmodel}
\begin{tabular}{llcccc}
\hline
\multirow{2}{*}{\bf Approach} & \multirow{2}{*}{\bf Model} & \multicolumn{2}{c}{\bf Dev} & \multicolumn{2}{c}{\bf Test} \\ \cline{3-6} 
 &  & {\bf EM} & {\bf F1-score} & {\bf EM} & {\bf F1-score} \\ \hline
\multirow{2}{*}{\bf Ranking} & Word Count & 68.14 &  73.69 & 67.47 & 73.21 \\ 
 & BM25 & 74.05 & 78.86 & 73.17 & 78.23 \\ \hline
\multirow{2}{*}{\bf Classification} & maLSTM & 30.28 & 42.04 & 33.85 & 45.60 \\
 & BiGRU & 52.12 & 60.60 & 54.30 & 62.68 \\ \hline 
\multirow{5}{*}{\bf Machine Reading Comprehension} & QANet & 74.79 & 80.70 & 74.84 & 81.20 \\ 
 & mBERT & 86.35 & 89.16 & 84.74 & 88.57 \\ 
 & PhoBERT & 83.33 & 85.13 & 79.91 & 81.93 \\ 
 & XLM-R$_{Base}$ & 83.14 & 87.04 & 82.97 & 87.10 \\ 
 & {\bf XLM-R$_{Large}$} & {\bf 91.79} & {\bf 93.95} & {\bf 85.87} & {\bf 88.77} \\ \hline 
\end{tabular}
\end{table}


\subsection{Result Analysis}

\begin{figure}[H]
    \centering
    \includegraphics[width=1\linewidth]{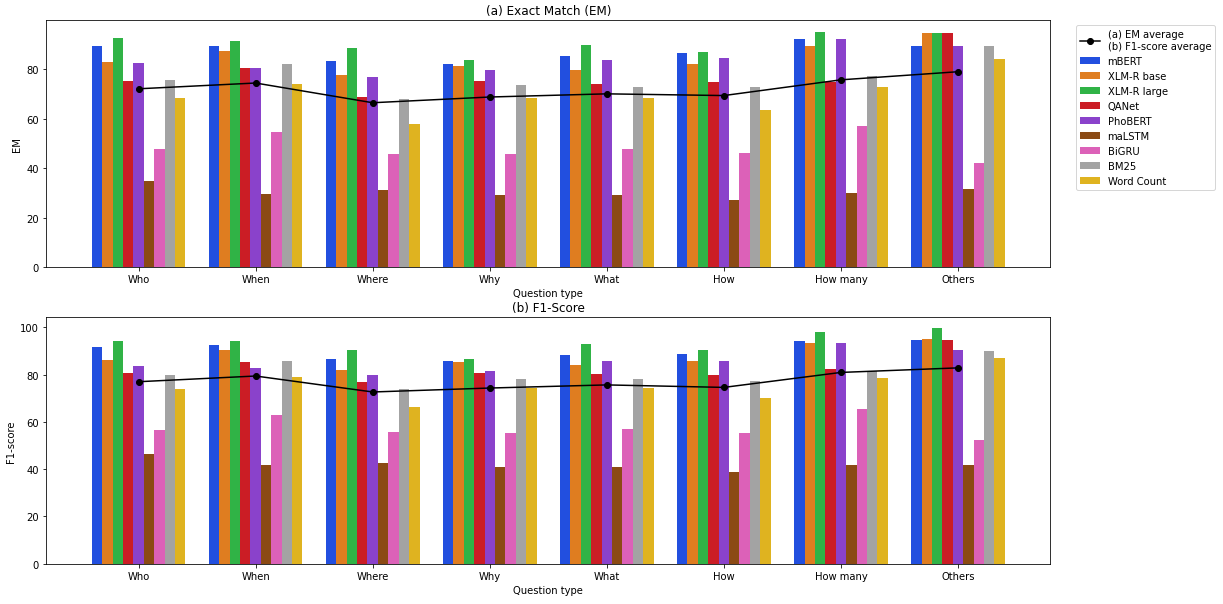}
    \caption{Performance of the models on each question type in UIT-ViWikiQA. The line graphs are the average performance of the models for each question type.}
    \label{fig:F1andEM_on_question_type}
\end{figure}

We measure the performance of each model for each question type using F1-score and EM on the Dev set of our dataset, and the results are shown in Figure \ref{fig:F1andEM_on_question_type}. In general, most models of the MRC-based approach outperform the rest in all question types. The second are models of the ranking-based approach, and the last is the classification-based approach models. Among them, the BERT models and their variants have the best performance. In particular, the XLM-R$_{Large}$ model has the highest F1-score and EM in all question types. The models score lower for the questions such as Where, Why, What, and How. We found that the amount of data, the variety of question words, and the difficulty of the questions affected the performance of the models. In particular, Where questions have a low quantity and a wide variety of words to pose questions (see Figure \ref{fig:question_types}), all models have a lower F1-score and EM in this question type. The What question has many data samples (49.06\%), but it has a very high diversity of words to ask. It makes it difficult for the models to predict answers. Why and How are difficult questions that require the computer to understand context and questions, which achieve low F1-score and EM scores. The question types where the question word to pose is not diverse have yielded high scores for models such as How many type questions. The analysis of experimental results is based on question types to evaluate the difficulty of questions for Vietnamese sentence extraction-based MRC on our data set, helping researchers have more ideas for the research in future.

\begin{table}[H]
\centering
\caption{Performance of MRC models on UIT-ViWikiQA and its shuffle version.}
\label{tab:compare_shuffle}
\begin{tabular}{ccccccccc}
\hline
\multirow{3}{*}{\textbf{Model}} & \multicolumn{4}{c}{\textbf{Original Version}}                                                                                                          & \multicolumn{4}{c}{\textbf{Shuffle Version}}                                                                                                  \\ \cline{2-9} 
                                & \multicolumn{2}{c}{\textbf{Dev}}                                        & \multicolumn{2}{c}{\textbf{Test}}                                       & \multicolumn{2}{c}{\textbf{Dev}}                                        & \multicolumn{2}{c}{\textbf{Test}}                                       \\ \cline{2-9} 
                                & \multicolumn{1}{c}{\textbf{EM}} & \multicolumn{1}{c}{\textbf{F1-score}} & \multicolumn{1}{c}{\textbf{EM}} & \multicolumn{1}{c}{\textbf{F1-score}} & \multicolumn{1}{c}{\textbf{EM}} & \multicolumn{1}{c}{\textbf{F1-score}} & \multicolumn{1}{c}{\textbf{EM}} & \multicolumn{1}{c}{\textbf{F1-score}} \\ \hline
\textbf{QANet}                  & 74.79                           & 80.70                                  & 74.84                           & 81.20                                  & 73.85                           & 80.19                                  & 73.84 (-1.00)                   & 80.48 (-0.72)                          \\
\textbf{PhoBERT}                & 83.33                           & 85.13                                  & 79.91                           & 81.93                                  & 81.88                           & 84.34                                  & 78.09 (-1.82)                   & 80.74 (-1.19)                          \\
\textbf{mBERT}                  & 86.35                           & 89.16                                  & 84.74                           & 88.57                                  & 83.80                           & 87.67                                  & 83.56 (-1.18)                   & 87.93 (-0.64)                 \\
\textbf{XLM-R$_{Base}$}             & 83.14                           & 87.04                                  & 82.97                           & 88.57                                  & 81.82                           & 86.13                                  & 81.34 (-1.63)                   & 85.43 (-3.14)                          \\
\textbf{XLM-R$_{Large}$}            & \textbf{91.79}                  & \textbf{93.95}                         & \textbf{85.87}                  & \textbf{88.77}                         & \textbf{90.30}                  & \textbf{92.91}                         & \textbf{84.25 (-0.79)}          & \textbf{87.98 (-1.62)}                          \\ \hline
\end{tabular}
\end{table}

Of the three approaches we propose in Section \ref{approachs}, the approach based on MRC has context attached to the input. We conduct experiments to demonstrate the dependence of the models of the MRC-based approach to the context, and Table \ref{tab:compare_shuffle} shows the result. We create another version of the UIT-ViWikiQA dataset by shuffling sentences in its context. The performance of all models decreased in the version with the shuffle context. All models have an average decrease of 1.45\% in EM and an average decrease of 1.00\% in F1-score on the Test set, which proves that context influences the performances of the models. The XLM-R$_{Large}$ model has the highest performance in both EM and F1-score in both versions of our dataset.



%% file: section/6-Conclusionandfuturework.tex
\section{Conclusion and Future Work}

In this paper, we introduced UIT-ViWikiQA, a new sentence-extraction dataset for evaluating Vietnamese MRC. This dataset comprises 23.074 question-answers based on 5.109 passages of 174 Vietnamese articles from Wikipedia. According to our experimental results when extracting the sentence, the XLM-R$_{Large}$ model achieved the highest performances on the Test set with 85.87\% of EM and 88.77\% of F1-score. On the contrary, when shuffling sentences to change context, the MRC models obtained an average decrease of 1.45\% in EM and an average decrease of 1.00\% in F1-score on the Test set, proving that context influences the performances of the models. Our result analysis explored the challenging questions to be addressed in further studies: Where, Why, What, and How.

In future, we propose several directions: (1) increasing the quantity of complex questions can boost the performance of models; (2) addressing the challenging questions using knowledge base \cite{yang2019enhancing} is to enhance the performance of this task; (3) researchers will exploit the challenging task of multilingual sentence extraction-based MRC; (4) extracting evidence sentences can help improve span-extraction reading comprehension on the multiple MRC datasets which are SQuAD \cite{rajpurkar2016squad}, UIT-ViQuAD \cite{nguyen-etal-2020-vietnamese}, FQuAD \cite{dhoffschmidt2020fquad}, SberQuAD \cite{Efimov_2020}, KorQuAD \cite{DBLP:journals/corr/abs-1909-07005}, SQuAD-IT \cite{SQuADit}, and CMRC \cite{cui-etal-2019-span}; (5) sentence answer-extraction question answering can be developed inspired by this study and DrQA \cite{chen2017reading}.
